\documentclass[times, review, 10pt]{elsarticle}

\usepackage{amssymb}
\usepackage{amsmath}

\usepackage{svg}
\usepackage{booktabs}   
\usepackage{multirow}   
\usepackage{diagbox} 
\usepackage{hyperref} 
\usepackage{cleveref} 
\crefname{figure}{Fig.}{Figs.} 
\crefname{table}{Table}{Tables} 

\journal{Pattern Recognition}

\begin{document}

\begin{frontmatter}



\title{CRIA: A Cross-View Interaction and Instance-Adapted Pre-training Framework for Generalizable EEG Representations}


\author[1,2,3]{Puchun~Liu} 
\author[1,2,3]{C. L. Philip~Chen}
\author[1,2,3]{Yubin~He}
\author[1,2,3]{Tong~Zhang\corref{cor1}}
\cortext[cor1]{The authors are with the Guangdong Provincial Key Laboratory of AI Large Model and Intelligent Cognition, the School of Computer Science and Engineering, South China University of Technology, Guangzhou 510006, China, and are with the Pazhou Lab, Guangzhou 510335, China, and are with Engineering Research Center of the Ministry of Education on Health Intelligent Perception and Paralleled Digital-Human, Guangzhou, China. (*Corresponding author: Tong~Zhang, E-mail: tony@scut.edu.cn). This work was funded in part by the National Natural Science Foundation of China grant under number 62222603, in part by the STI2030-Major Projects grant from the Ministry of Science and Technology of the People’s Republic of China under number 2021ZD0200700, in part by the Key-Area Research and Development Program of Guangdong Province under number 2023B0303030001, in part by the Program for Guangdong Introducing Innovative and Entrepreneurial Teams (2019ZT08X214), and in part by the Science and Technology Program of Guangzhou under number 2024A04J6310.}

\affiliation[1]{
  organization={Guangdong Provincial Key Laboratory of AI Large Model and Intelligent Cognition, School of Computer Science and Engineering, South China University of Technology},
  addressline={382 Waihuan Rd E},
  city={Guangzhou},
  postcode={510006},
  state={Guangdong},
  country={China}
}
\affiliation[2]{
  organization={Pazhou Lab},
  addressline={248 Qiaotou Street},
  city={Guangzhou},
  postcode={510006},
  state={Guangdong},
  country={China}
}
\affiliation[3]{
  organization={Engineering Research Center of the Ministry of Education on Health Intelligent Perception and Paralleled Digital-Human},
  addressline={382 Waihuan Rd E},
  city={Guangzhou},
  postcode={510006},
  state={Guangdong},
  country={China}
}
\begin{abstract}
The difficulty of extracting deep features from EEG data and effectively integrating information from multiple views presents significant challenges for developing a generalizable pretraining framework for EEG representation learning. However, most existing pre-training methods rely solely on the contextual semantics of a single view, failing to capture the complex and synergistic interactions among different perspectives, limiting the expressiveness and generalization of learned representations. To address these issues, this paper proposes CRIA, an adaptive framework that utilizes variable-length and variable-channel coding to achieve a unified representation of EEG data across different datasets. In this work, we define cross-view information as the integrated representation that emerges from the interaction among temporal, spectral, and spatial views of EEG signals. The model employs a cross-attention mechanism to fuse temporal, spectral, and spatial features effectively, and combines an attention matrix masking strategy based on the information bottleneck principle with a novel viewpoint masking pre-training scheme. Experimental results on the Temple University EEG corpus and the CHB-MIT dataset show that CRIA outperforms existing methods with the same pre-training conditions, achieving a balanced accuracy of 57.02\% for multi-class event classification and 80.03\% for anomaly detection, highlighting its strong generalization ability.
\end{abstract}




\begin{keyword}


EEG Representation \sep Cross-view learning \sep Instance adaptive framework \sep Representation purification
\end{keyword}

\end{frontmatter}


\section{Introduction}
Electroencephalography (EEG) is a non-invasive method for recording real-time cerebral electrical activity, offering high temporal resolution and enabling long-term monitoring in neuroscience, medical diagnostics, and brain-computer interfaces. Due to its ability to characterize deep brain functional responses, EEG has become essential in fields such as epilepsy \cite{jing2023development, arroyo1992high}, sleep staging \cite{taran2020automatic, jia2020graphsleepnet, zhang2025psss}, driver drowsiness recognition \cite{tang2025riding}, affective computing \cite{murugappan2013human, song2018eeg, zhang2019gcb, zhang2022visual}, and cognitive science \cite{werkle2006cortical}. Pre-training models such as BERT \cite{devlin2018bert} and GPT \cite{achiam2023gpt} have revolutionized the fields of computer vision (CV) and natural language processing (NLP), enhancing downstream task performance by learning generic representations from vast amounts of unlabeled data. These self-supervised techniques have also been successfully applied to EEG analysis. Wang et al.'s BrainBERT \cite{wang2023brainbert} discretizes EEG data into “Brain Words” and uses a BERT-like mask prediction task for pre-training. Yang et al.'s BIOT method \cite{yang2024biot}, on the other hand, combines frequency-domain features and perturbation pretraining strategies. Larger-scale EEG modeling frameworks, such as Brant \cite{zhang2024brant, yuan2024brant} by Zhang et al. and LaBraM \cite{jiang2024large} by Jiang et al., employ large-scale datasets for multi-task pre-training and reduce the high variability and noise of the data by vectorizing the EEG data, thereby improving model performance.

However, EEG data are highly non-stationary and noisy, making traditional pre-training methods less effective \cite{bhatti2024comparative}. Brant \cite{zhang2024brant, yuan2024brant} mitigates this issue with purer SEEG data, but complicates data acquisition. In contrast, LaBraM \cite{jiang2024large} reduces noise via VQVAE, albeit at the expense of increased training time. In addition, different views of EEG data contain complementary information, and previous methods basically analyzed them from a single-view perspective or used a multi-view serial structure, which does not fully utilize the parallel mutual information between views. Finally, the problem of data heterogeneity has not been effectively addressed due to the diverse acquisition methods and experimental paradigms used in EEG datasets. To address the above challenges, the primary goal of this study is to develop a pre-training framework that can effectively handle the multi-view characteristics of EEG data, adapt to data heterogeneity, and enhance the model's generalization ability.

This paper introduces the CRIA (Cross-View instance-adaptive Pre-training model) framework, which aims to enhance the characterization learning of EEG data through multi-view interaction and adaptive processing. CRIA integrates a cross-attention-based Multi-view Feature Fusion Mechanism and instance-adaptive variable-length/channel encoding to enhance temporal, spatial, and frequency features of EEG data, enabling unified multi-dataset modeling beyond single-view limitations. It introduces a View-wise Masking Pre-training Strategy and an attention masking method grounded in information bottleneck theory, reducing noise impact in self-supervised learning, improving adaptability to non-stationary EEG, and enhancing generalization while lowering overfitting risk. Additionally, its Representation Enhanced Purification Strategy exploits variations in EEG event salience across time and channels to emphasize task-relevant information. Overall, CRIA provides a flexible, general and high-performance pre-training framework for EEG data analysis.

The main contribution of this paper includes:
\begin{enumerate}
    \item This paper presents CRIA, a novel asymmetric three-view interaction pre-training framework for EEG analysis, in which the spectral view is designated as the dominant modality and lateral cross-attention is performed from temporal and spatial views. As the first EEG pre-training strategy to model view fusion from a dominant–auxiliary modality perspective, CRIA effectively captures complementary inter-view information while preserving view-specific features. By integrating such structured cross-view interactions, CRIA enhances EEG representation learning and ensures a smooth transition from pre-training to downstream fine-tuning, thereby improving model performance and generalization.
    \item CRIA adopts an innovative view masking pretraining strategy to enhance cross-view perceptual inference, overcoming the limitations of traditional context masking. It utilizes a cross-attention module to focus on cross-view interactions within EEG data, thereby strengthening its representation across temporal, spatial, and frequency views. The enhanced purification strategy addresses insufficient attention to EEG event differences across time periods and channels. During fine-tuning, the attention matrix mask method improves stability, generalization, and reduces overfitting. This strategy is validated through information bottleneck theory.
    \item Benefiting from the design of the pre-training framework to downstream task fine-tuning, and the innovative view mask pre-training strategy, CRIA is able to realize instance adaptive coding to flexibly process EEG data with different lengths and numbers of channels. This design enables CRIA to provide flexible processing paths across different datasets and tasks.
\end{enumerate}

Experimental results demonstrate that CRIA outperforms existing methods in tasks such as abnormal EEG detection and multi-class EEG event classification. In addition, CRIA still exhibits stronger robustness in the absence of pre-training conditions and with the addition of various types of noise.

\section{Related Work}
EEG research has evolved from traditional signal processing to modern deep learning and pre-training strategies. Time, frequency, and wavelet analyses have laid the groundwork for detecting neurological disorders. Deep learning, including convolutional, recurrent, and graph-based models, captures complex patterns, while multi-modal and multi-view learning enhance robustness. Recently, self-supervised pre-training has leveraged large-scale unlabeled data, enabling the development of generalizable, high-performing models. Representative work in related areas will be presented in this chapter.

\subsection{Signal processing-based EEG analysis}
Among the various methods, time-domain analysis remains one of the most fundamental techniques, which emphasizes the temporal characteristics of EEG data. This category of analysis encompasses a range of approaches that examine how EEG signals change over time, revealing neural responses to specific events by analyzing features such as amplitude, latency, and waveform. The sensitivity of time-domain analysis to these features makes it useful in identifying neurological disorders\cite{da2013eeg}. 

Time-domain methods offer direct insights into EEG dynamics but struggle with noise reduction and information compression. To mitigate this issue, frequency-domain approaches, including the fast Fourier transform (FFT), emphasize spectral analysis and have demonstrated success in tasks such as emotion recognition \cite{murugappan2013human} and seizure detection. However, the FFT only provides global frequency information, which limits its ability to capture transient features in EEG data. The short-time Fourier transform (STFT) improves on this by enabling localized frequency analysis through a sliding window, helping detect temporal variations in nonstationary signals \cite{allen1977short}. Despite its benefits, the fixed window size of the STFT results in a trade-off between time and frequency resolution, limiting its adaptability for complex temporal variations.

To address these challenges, the wavelet transform has emerged as a versatile alternative method, providing a more flexible framework for time-frequency analysis. Unlike Fourier-based methods, the wavelet transform offers multiresolution analysis, enabling adaptive resolution in both the time and frequency domains. Daubechies' pioneering work \cite{daubechies1990wavelet} introduced compactly supported wavelets, laying the theoretical foundations for this method. Since then, the wavelet transform has been widely used in EEG data analysis, including epilepsy detection \cite{alturki2020eeg} and sleep staging \cite{taran2020automatic}. There is also some work that uses deep clustering to address overfitting in small sample scenarios \cite{tabejamaat2025eeg}.

\subsection{Deep learning for EEG tasks}
Deep learning has significantly advanced EEG analysis by extracting deep, task-relevant information. Schirrmeister et al. \cite{schirrmeister2017deep} proposed DeepConvNet, a CNN-based model that effectively decodes EEG motion images by capturing spatial and temporal features, paving the way for CNN applications in EEG. Recurrent neural networks, such as LSTMs, have also shown success in modeling long-term dependencies and dynamic patterns \cite{zheng2020ensemble}, particularly in EEG emotion recognition tasks \cite{craik2019deep}. Zhang et al. \cite{zhang2019gcb} expanded on Graph neural networks (GNNs) with GCB-Net, which combines GCN and BLS to enhance EEG emotion classification. However, these models suffer from long-range forgetting, limited inter-electrode correlation modeling, poor generalizability, and high computational demands on large-scale EEG data.


Attentional mechanisms further improve EEG analysis by allowing models to focus on key temporal and spatial features. Jia et al. \cite{jia2020graphsleepnet} introduced GraphSleepNet, which combines CNNs with self-attention for superior pattern recognition in EEG data, achieving state-of-the-art performance in sleep stage classification. Yi et al. \cite{ding2023lggnet} employ attentional mechanisms to investigate the cooperative associations between different brain regions, modeling the complex relationships within and between functional brain areas. These developments mark the shift from basic CNNs and RNNs to advanced graph-based and attention-enhanced models, improving accuracy and interpretability.

\subsection{Multi-modal learning and multi-view learning}
Multi-modal learning combines information from different data acquisition methods (e.g., EEG, fMRI, NIRS), while multi-view learning extracts complementary information from various perspectives of a single data modality. These approaches often complement each other in practice. Cross-attention is frequently used to fuse different modalities, views, and scales \cite{chen2021crossvit}. Cross-attention is commonly used in EEG analysis to integrate information from various input sources, such as spatial regions or different feature representations. For instance, Pahuja et al.'s XAnet \cite{pahuja2023xanet} uses cross-attention to integrate left and right brain regions. 

EEG researchers are often concerned with multi-view model effects, as seen in the Multi-View Spectral-Spatial-Temporal Masked Autoencoder (MV-SSTMA) \cite{gao2024multimodal}. MV-SSTMA uses a hybrid encoder-decoder with multi-scale causal convolutions and multi-head self-attention to extract variable-length features, followed by spatial self-attention for cross-scale fusion. It emphasizes robust three-view serial reconstruction and fusion of EEG and eye movement signals to enhance emotion recognition. In contrast, CRIA focuses on cross-attention-based multi-view parallel interaction, adaptive coding across EEG datasets, and a unique view-mask pre-training strategy. Notably, MV-SSTMA extracts temporal features from frequency inputs and applies spatial self-attention, but it is not true multi-view learning. CRIA and MV-SSTMA offer complementary, non-conflicting innovations that can be jointly applied. In addition, the MCSP \cite{wei2025multi} proposed by Wei et al. employs cross-domain self-supervised loss and introduces cross-modal self-supervised loss, which utilizes the complementary information of fMRI and EEG to facilitate in-domain knowledge refinement and maximize cross-modal feature convergence. This is similar to CRIA in fusing multi-view information, but the focus is on balancing multiple losses. The CBraMod model \cite{wang2024cbramod} proposed by Wang et al. changes the traditional attention mechanism into criss-cross attention, which exploits the structural features of EEG signals to simulate spatial and temporal dependencies through two parallel modalities. The core innovation lies in the unique attention structure that captures the spatial and temporal dependencies of EEG.

Compared to other cross-modal attention mechanism models, which mainly use stacked Transformer layers for information fusion and lack of fine-grained modeling of modal independence and complementarity, the main innovation of CRIA is that it is the first time to construct an asymmetric three-view lateral interaction pre-training framework with primary and secondary modalities in the field of EEG analysis, which models EEG signals from the view fusion and enhancement perspectives. In addition, CRIA's instance-adaptive feature and unique viewmask pretraining strategy can also reflect its innovation.

\subsection{EEG self-supervised pre-training}
Self-supervised pre-training improves model performance on downstream tasks by learning generic feature representations from large amounts of unlabeled EEG data. The BrainBERT model \cite{wang2023brainbert} discretizes EEG data into ``Brain Words'' and uses a BERT-like mask prediction task for pre-training. However, it is limited in incorporating multi-view and multi-domain interactions. The BIOT method \cite{yang2024biot} enhances flexibility by utilizing frequency-domain features and perturbation-based pre-training; however, its reliance on frequency data alone may overlook temporal and spatial dynamics. Zhang et al. \cite{zhang2024brant, yuan2024brant} proposed Brant, which utilizes large-scale, self-collected SEEG and EEG data for multi-task pre-training, with models reaching up to 500M parameters. While Brant benefits from a data-driven approach, its focus on multi-task learning and model size doesn’t fully address EEG data’s heterogeneity and variability across datasets. Additionally, the high cost of data collection and labeling hinders research progress. Another notable advancement in pure EEG self-supervised learning is the Large Brain Model (LaBraM) \cite{jiang2024large}, which explores masked modeling with a two-stage pre-training approach by VQVAE. Although LaBraM effectively addresses EEG noise and variability through vectorization, it still faces the challenge of balancing the trade-off between time-frequency resolution in EEG data analysis. Additionally, large-scale pre-training of up to 2500 hours can be computationally expensive and may not be feasible for all applications. 

A common problem faced by these methods is the inability to fully utilize the information from parallel interactions of multiple viewpoints, as well as their susceptibility to the high variability, non-stationarity, and data heterogeneity of EEG data. In contrast, CRIA fuses multi-view features in temporal, frequency and spatial views through a cross-view instance adaptive pre-training framework, and employs a unique view-wise masking pre-training strategy to deal with the pre-training convergence problem of EEG data. CRIA improves the adaptability to non-smooth EEG data and enhances the generalization ability while reducing the risk of overfitting. In the latter part of the paper, we will quantitatively verify the advantages of CRIA over other models through detailed comparative experiments.

\begin{figure*}[t]
\centering
\includegraphics[width=1\textwidth]{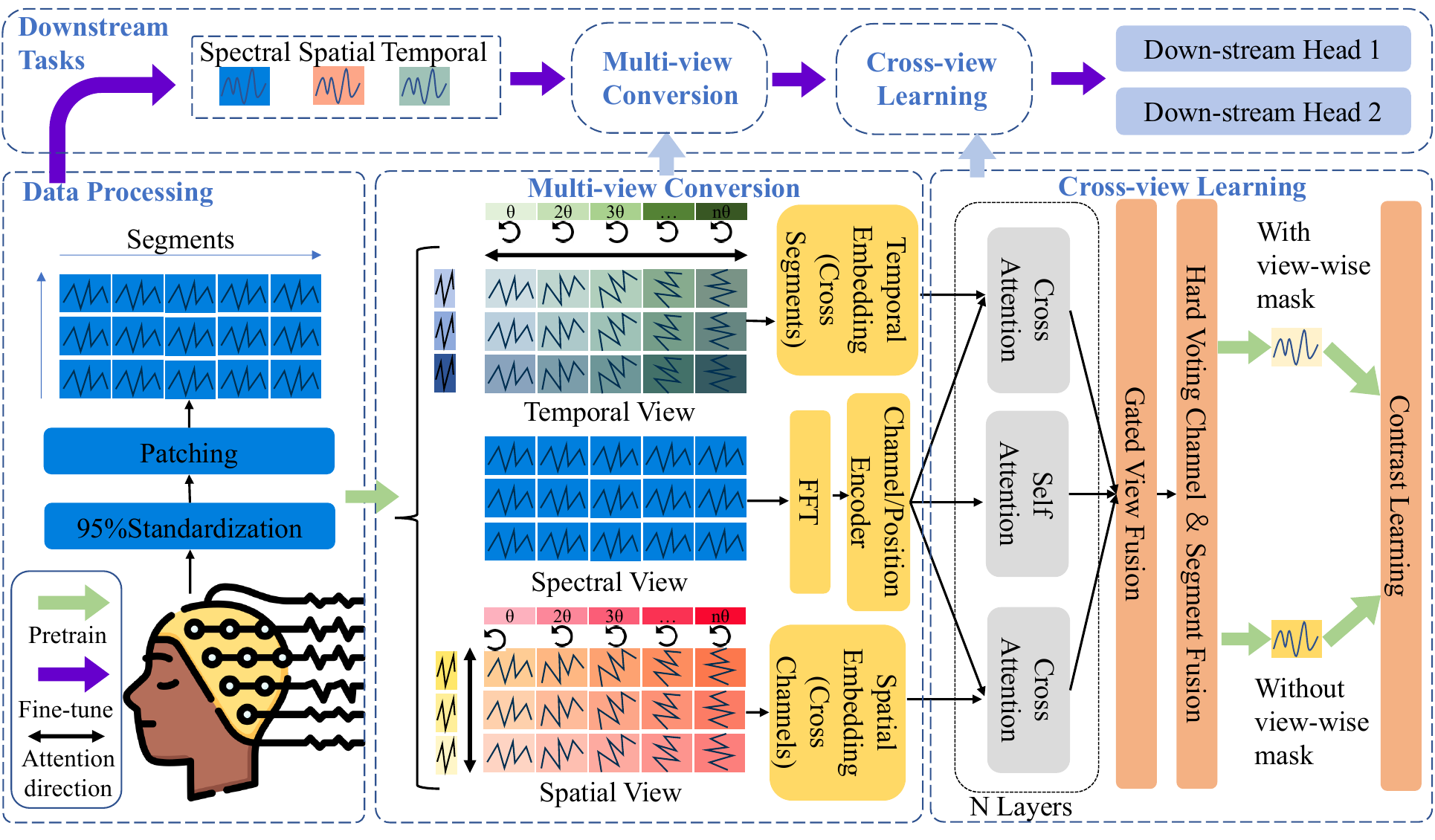}
\caption{CRIA's overall framework structure. The top box represents the downstream fine-tuning, and the bottom box represents the view-wise mask pre-training strategy.}
\label{model}
\end{figure*}

\section{Methodology}
The overall structure of CRIA is shown in Fig.~\ref{model}, which aims to propagate information between views and reinforce the richness of traditional attention features by paying cross-attention to multiple views of EEG data. EEG data are multi-channel time series data, which can be represented as $\mathbf{S} \in \mathbb{R}^{C \times L_o}$ after being split into slices under different task settings, acquisition conditions and events. Here, $\mathbf{S}$ denotes the EEG data, $C$ represents the number of channels, and $L_o$ indicates the number of sampling points at the original sampling rate. CRIA is designed to adapt to EEG data of different lengths or channels, enabling self-supervised pre-training on them, which benefits further downstream tasks. In the subsequent subsections, we present the complete modeling framework, including Dataset Pre-Processing, Multi-view Conversion and Position Encoding, View Interaction Feature Enhancement by Cross-attention, Feature Fusion and Representation Enhanced Purification, as well as Pre-training and downstream tasks.

\subsection{Dataset Pre-processing}
\subsubsection{Resampling}
Studies have shown that the vast majority of EEG activity examined during clinical EEG interpretation is below 30 Hz \cite{gotman2010high}, but signals as high as 120 Hz can be seen in EEG studies during seizures \cite{arroyo1992high}. In addition, short-wave regions in the 100-200 Hz range are often seen as EEG ``ripples'' associated with normal physiological activity \cite{staba2004high}. In this work, EEG data is resampled to 200 Hz for semantic alignment in the frequency domain. 
\subsubsection{Filtering}
The resampled data is pre-processed in multiple steps and corresponds to the work being compared: First, a Butterworth band-pass filter (0.5–120 Hz) is used to retain the major EEG frequency components, then an IIR trap filter (1 Hz, 60 Hz) removes drift and interference. Finally, the dataset was sliced into varying lengths based on downstream tasks, with resampled lengths from $L_o$ to $L$. To prevent excessive differences in EEG data amplitude across subjects and trials, all data were normalized to the 95th percentile.

\subsection{Multi-view Conversion}
An EEG data slice $S \in \mathbb{R}^{C \times L}$ can be segmented in the time dimension like $S' \in \mathbb{R}^{C \times N \times D}$, where $N$ is the number of segments, and $D$ is the length of each segment which is also the Transformer hidden layer dimension. The model in this paper extracts interactive features from temporal, spectral, and spatial views, enabling a comprehensive representation of the data. Traditional methods rely on either manually crafted or machine-learned features: manual features are interpretable and efficient but lack generalizability, while learned features capture complex patterns yet may be unstable and sensitive to noise. We extract temporal and spatial features using a single layer of the Linear Attention Transformer (LT) \cite{wang2020linformer}, with rotary position encoding (RoPE) \cite{su2024roformer} applied to the temporal dimension beforehand to ensure that the attention mechanism accounts for the order of the input signal. Both the Linear Attention Transformer and the rotary position encoding are able to act on different number of segments N within the maximum length, a property suitable for use in pre-training to cope with different lengths of sample data and cut-off paradigms.

Suppose the data for a channel in $S'$ is $\mathbf{X_c} = (x_1, x_2, ... , x_{D}) \in \mathbb{R}^{N \times D}$, where $N$ denotes the number of time segments and $D$ denotes the hidden layer dimension. For the purpose of the rotation operation, we consider each row of data in $\mathbf{X_c}$ as a combination of $D/2$ complex values:
\begin{align}
\begin{split}
    \mathbf{X_c} &= (x_1 + ix_2, x_3 + ix_4, ... , x_{D-1} + ix_{D}) \\
                 &= (z_1, z_2, ... , z_{D/2}),
\end{split}
\end{align}
where $n \in [1, N]$ is the time segment index and $d \in [1, D/2]$ is the index that represents the complex dimension. Specifically, for each $d$, the complex-valued representation is formed as $z_d = x_{2d-1} + ix_{2d}$, where $x_{2d-1}$ and $x_{2d}$ denote the real and imaginary parts. Thus, the complex index $d$ refers to the position in the resulting complex-valued sequence of length $D/2$. For time segment $n$ and complex index $d$, we define the rotation angle to be:
\begin{align}
    \theta_{n,d} = n \cdot 10000^{-\frac{2d}{D}},
\end{align}
where $10000^{-\frac{2d}{D}}$ is a frequency scaling factor that ensures that different dimensions are encoded with different frequencies, and $n$ is used to introduce positional information at the time step. Based on the rotation angle, the complex form of the RoPE operation is:
\begin{align}
    \text{RoPE}(\mathbf{X_c}, n, d) = (x_{2d-1}+ix_{2d}) \cdot e^{i\theta_{n,d}},
\end{align}
\begin{align}
    e^{i \theta_{n, d}}=\cos \left(\theta_{n, d}\right)+i \sin \left(\theta_{n, d}\right),
\end{align}
where $e^{i \theta_{n, d}}$ stands for complex rotation. After applying this formula to all time steps $n$ and all complex dimensions $d$, we obtain the matrix processed by rotational position encoding:
\begin{align}
    S_{r,v} = \text{RoPE}(S'_v),
\end{align}
where $S_{r,v}$ means slice afer rotation in view $v$, and $v \in \{temporal,spatial\}$.

To fully utilize the temporal information introduced by the position encoding and efficiently capture long-range dependencies and feature interactions between sequences, we add an LT layer after the position encoding. The complexity of the standard self-attention mechanism is $O(N^2)$, whereas the linear attention mechanism reduces this complexity by introducing kernelization, resulting in $O(N)$. Assuming that the optional kernel function is $\phi(\cdot)$, Q and K can be projected into a higher dimensional space, making the inner product exchangeable:
\begin{equation}
    \text {LT}(\mathbf{Q}, \mathbf{K}, \mathbf{V})=\phi(\mathbf{Q})\left(\phi(\mathbf{K})^{\top} \mathbf{V}\right).
\end{equation}
This avoids explicitly storing $\mathbf{QK}^{\top}$ in the attention mechanism, dramatically reducing the memory footprint.

In order to achieve uniform modeling of datasets with different number of channels, a learnable $E_{channel} \in \mathbb{R}^{C_{max} \times D}$ is maintained to label channels in the attention mechanism, where $C_{max} $ is the maximum number of channels in the pre-training dataset. During pre-training, $C_{max} $ contains all occurrences of EEG channels, although some of the channels represent a small percentage of the overall sample space. The model automatically matches the encoding of the channel corresponding to the data during forward propagation, which is used to indicate the channel information in the attention mechanism, allowing the model to learn the roles of the different channels and apply it to all samples of channel combinations. Then the final feature before entering the cross-attention layer is: 
\begin{equation}
    F_{tem} = \text{LT}(S_{r,temporal} + E_{channel}[1:C]),
\end{equation}
\begin{equation}
    F_{spa} = \text{LT}(S_{r,spatial} + E_{channel}[1:C]).
\end{equation}
In EEG data analysis, the spatial view is typically represented by channel information. Consequently, instead of being entirely flattened, the data are input into the LT along two dimensions: the channel dimension and the segment dimension, which serve as the spatial and temporal feature views.

In practice, it has been found that there is a need for a view that retains the more primitive state without extracting features through a deeper network, which would otherwise cause the model to forget previously learned capabilities in case of data malfunction. To address this problem, we retain stable and low-noise frequency view information in the view to prevent training instability and corrupted spectral features. Specifically, the method utilizes a one-dimensional Fast Fourier Transform (FFT) to extract the amplitude information of the signal, which is then combined with rotational position coding to augment the frequency features. Compared to the other two views, it inserts the position codes in a different order:
\begin{equation}
    F_{spe} = \text{RoPE}(\text{FFT}(S'_{spectral} ) + E_{channel}[1:C]).
\end{equation}
Fast Fourier Transform's dimensionality can be aligned with temporal and spatial features, and can be applied equally well to all combinations of sample lengths and channels. In summary, the entire Multi-view Conversion stage model can receive pre-trained data of variable lengths and channels as input.

\subsection{Cross-Attention Learning with Representation Purification}
\subsubsection{Cross-attention Feature Enhancement}
Previous methods have input different channels and time slices of EEG data into the attention mechanism, treating them as if they are the same view, or have used only one of them \cite{yang2024biot, chen2024eegformer}. Some methods use a two-step approach for attention computation \cite{zhang2024brant, yuan2024brant}. However, this can destroy the feature information of a single view or amplify noise within the attention mechanism. Inspired by multimodal feature fusion \cite{chen2021crossvit}, we propose a method that utilizes cross-attention to attend to multiple views of the EEG data in parallel. Choosing the spectrum as the primary view helps ensure that at least one view remains stable and is not excessively influenced by the other two during multi-view learning. Self-Attention is computed by weighting and summing all input elements based on the relationship between each element of the input, thus providing a weighted contextual information for the output at each position:
\begin{equation}
    \mathrm{SA}(F)=\operatorname{softmax}\left(\frac{Q K^T}{\sqrt{d_k}}\right) V,
\end{equation}
\begin{equation}
    Q=FW_Q,K=FW_K,V=FW_V,
\end{equation}
where $d_k$ is the dimension of the Key vector, which is used to scale the dot product so that it is not too large to cause the gradient to vanish or explode. $Q$, $K$, and $V$ are all computed at the same frequency view feature matrix. $W_Q$, $W_K$ and $W_K$ are learnable matrices. In practice, Multi-Head Attention (MHA) are often used to enhance the representation of a model:
\begin{equation}
    \text { MHA }(Q, K, V)=\text { Concat }\left(\operatorname{head}_1, \ldots, \operatorname{head}_h\right) W_O,
\end{equation}
\begin{equation}
    \operatorname{head}_i=\operatorname{SA}\left(Q W_Q^i, K W_K^i, V W_V^i\right),
\end{equation}
\begin{figure}[t]
\centering
\includegraphics[width=0.6\columnwidth]{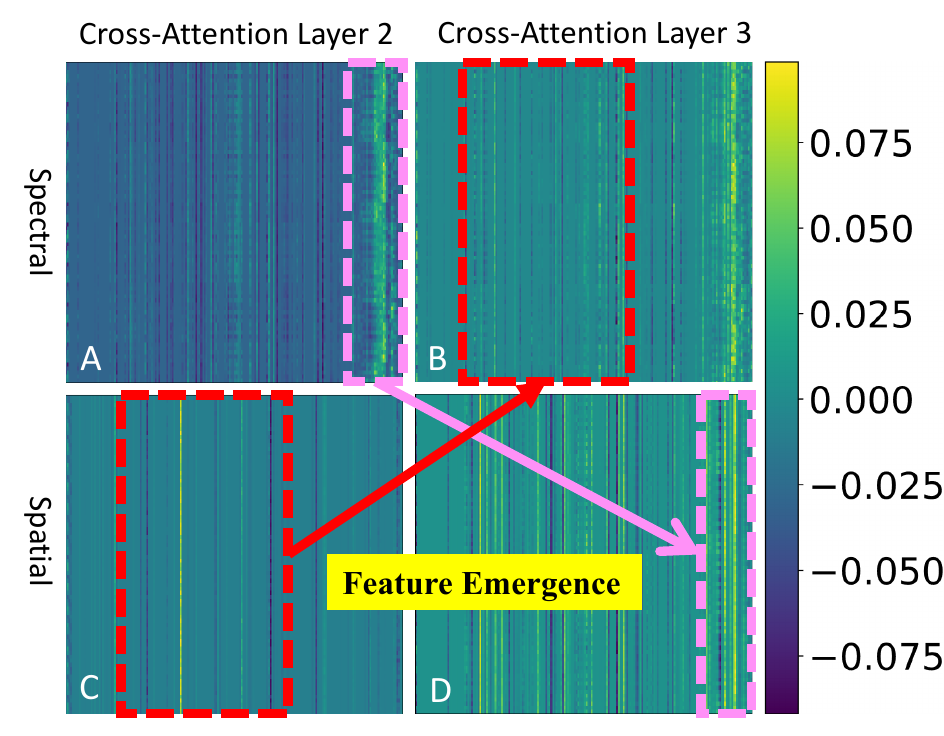} 
\caption{Feature visualizations of frequency and spatial views across different layers of the Cross-Attention module. Subfigures A and B show the spectral view at Cross-Attention Layers 2 and 3. Subfigures C and D correspond to the spatial view at Layers 2 and 3. The dashed boxes highlight regions where distinct feature activations emerge, with red indicating spatial patterns and magenta indicating spectrum-salient regions. }
\label{FeatureCrossAttention}
\end{figure}

where $W_O$ is the linear transformation matrix of the output, $h$ is the number of heads, and each head computes the attention mechanism independently. Assuming that $l$ represents the current layer of the computation, the frequency view feature is calculated as:
\begin{equation}
    F_{spe,l}=\text{SA}(F_{spe,l-1}).
\end{equation}

In the same layer, temporal and spatial features are then considered as queries to participate in cross-attention (CA) in the frequency view. The calculation of cross-attention is similar to self-attention: 
\begin{equation}
    \mathrm{CA}(Q_{v_1}, K_{v_2}, V_{v_2})=\operatorname{softmax}\left(\frac{Q_{v_1} K_{v_2}^T}{\sqrt{d_k}}\right) V_{v_2},
\end{equation}
The $v_1$ and $v_2$ in the formula represent different views. Temporal and spatial features act as Queries in the attention mechanism, with spectral features as Keys and Values. Since different views provide complementary information, relying on a single view may cause information loss or incomplete feature representation. Simple feature fusion often fails to capture deep interactions. Cross-attention enables richer feature integration across views, mitigating these limitations. Moreover, due to its variable-length adaptability, both cross-attention (CA) and self-attention (SA) can process diverse pre-training data from the Multi-view Conversion stage. The key difference lies in CA using Queries and Key-Value pairs from different views of the same data—e.g., in EEG, temporal views reflect short-term signal changes, while spatial views capture inter-regional brain connectivity. Enhancements to frequency information from temporal and spatial views are detailed below:
\begin{equation}
    F_{spa,l}=\text{CA}(F_{spa,l-1},F_{spe,l-1},F_{spe,l-1}),
\end{equation}
\begin{equation}
    F_{tem,l}=\text{CA}(F_{tem,l-1},F_{spe,l-1},F_{spe,l-1}).
\end{equation}

Fig.~\ref{FeatureCrossAttention} illustrates the evolution of frequency and spatial features across different layers within the Cross-Attention module during forward propagation. Subfigures A and B depict the spectral view at Layer 2 and Layer 3, respectively, while subfigures C and D correspond to the spatial view at the same layers. The bold arrows in the figure indicate the direction of information flow between views, highlighting how spectral and spatial features interact and progressively align across layers. Notably, the saliency of features in one view is gradually enhanced through attention-based interactions with the complementary view, eventually manifesting in the other view—a phenomenon we refer to as ``Cross-view Feature Emergence”. For example, the vertically striped high-activation region within the red dashed box in the spatial view becomes evident in the frequency view of the next layer, while the granular oscillatory patterns in the pink dashed box in the frequency view are later reflected in the spatial representation. This cross-view reinforcement is particularly prominent in the highlighted regions, where similar activation patterns emerge across three views. These visualizations demonstrate that the Cross-Attention mechanism effectively facilitates feature exchange and mutual reinforcement between views, resulting in a gradual fusion of complementary information. This multi-view integration ultimately leads to richer, more coherent representations, enhancing the model’s overall discriminative capability.

\begin{figure}[t]
\centering
\includegraphics[width=0.6\columnwidth]{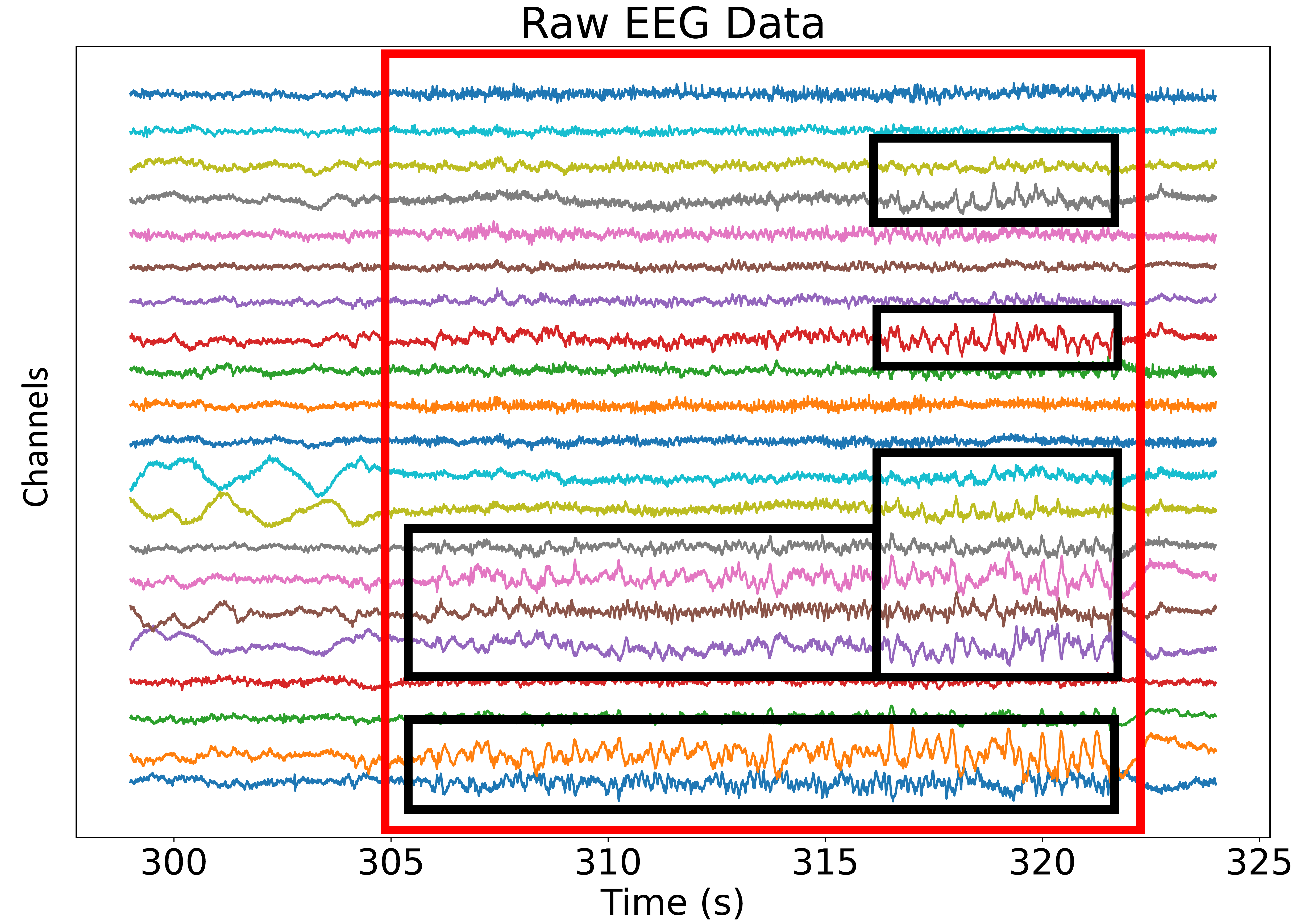}
\caption{Spatiotemporal regions exhibiting significant fluctuations in a given EEG event labeled by a human neurologist. The red box shows the overall time period of the event, and the black box shows the affected area.}
\label{RawEEG}
\end{figure}

\subsubsection{Representation Enhanced Purification}
We have observed that different regions of the EEG data respond in varying degrees to specific events, as shown in Fig.~\ref{RawEEG}. Certain channels (corresponding to specific brain regions) may show stronger or more consistent responses. Additionally, event-related potentials (ERPs) may exhibit distinct behavior in varying time windows. To make the model more focused on salient sites, we refined the feature fusion method using average mean pooling \cite{yang2024biot, zhang2024brant, yuan2024brant}. 

Assume that $C_{k_c}$ and $T_{k_t}$ are filtered for $ top k_c $ importance on all channels, followed by $ top k_t $ selection on the segment for each filtered channel. Finally, the features are constrained by applying an additional LayerNorm:
\begin{equation}
    C_{k_c} = \text{topk}\left(\frac{1}{N}\sum_{n=1}^T \left|\mathbf{x}_{c,n}\right|2, k_c\right),
\end{equation}
\begin{equation}
    T_{k_t} = \{ \text{topk}(\left|\mathbf{x}_{c,n}\right|2, k_t)| c \in C_{k_c} \},
\end{equation}
\begin{equation}
    F = \text{LayerNorm}\left(\frac{1}{C_{k_c}}\sum_{c\in C_{k_c}}(\frac{1}{T_{c,k_t}}\sum_{t\in T_{c,k_t}}\mathbf{x}_{c,n})\right).
    \label{eq:calculateF}
\end{equation}

\subsection{Pre-training and downstream tasks}
To enhance model expressiveness and robustness, we implemented a contrastive pre-training approach on random parameters. Considering the highly non-stationary nature of the EEG data, the reconstruction mask coding \cite{chen2024eegformer} or perturbation contrast learning \cite{yang2024biot} are not applicable. To solve this issue, we developed a straightforward pre-training method: for each sample, one of the three views was randomly masked and replaced with a placeholder. The model is trained to infer the missing-view information by leveraging the remaining two unmasked views, promoting the ability to reason across complementary modalities. We then computed the sample embeddings for both the unmasked and randomly masked conditions, followed by the calculation of the contrastive loss. This allows the model to pre-learn on tasks that are not overly difficult. To accomplish the masking of different views, three learnable auxiliary padding blocks for each view are first predefined: $A_{tem}$, $A_{spe}$, and $A_{spa} \in \mathbb{R}^{D}$. 

During forward propagation, a view is randomly selected for replacement. If the spectral view is chosen, the query, key, and value in the self-attention layer are replaced with $A_{spe}$; if the temporal or spatial view is chosen, the query in the cross-attention layer is replaced with $A_{tem}$ or $A_{spa}$, while the key and value remain unchanged. After view masking, a random view is hidden, prompting the model to rely on the remaining views for its predictions. This approach allows the model to learn deeper associations between views. Assuming that the output features after view masking are called $F^{\prime}$ and the regular output features without view masking are called $F$, which are obtained by a parameter-sharing model structure, contrastive loss\cite{he2020momentum} is used for pre-training of contrast learning:
\begin{equation}
    \mathcal{L} = \text{CrossEntropyLoss}\left(\text{softmax}\left(\langle F, F^\prime \rangle / T\right), \mathbf{I}\right).
\end{equation}
Here, \textit{T} represents the temperature (\textit{T} = 0.2 throughout the paper) and I is an identity matrix.

In the downstream task, CRIA uses a pre-trained encoder and a randomly initialized classification header. To weaken overfitting of the model due to excessive attention, we add a certain percentage of randomized attention value masking to all attention blocks, inspired by the information bottleneck (IB) \cite{saxe2019information}. Suppose $\theta$ is the set of parameters of the model, the formula for the information bottleneck is as follows:
\begin{equation}
    \max_{\theta} [I(A; Y) - \beta I(A; X)],
\end{equation}
where $A$ is the original attention matrix, $X$ and $Y$ are the data and label of the sample. Assuming that $M$ is a random mask matrix (with elements 0 or 1), then $A' = A \odot M$. We aim to show that the change in A after masking is consistent with the derivation of IB. According to the law of information chaining:
\begin{equation}
\begin{split}
    I(A, M ; X) &= I(A; X) + I(M; X \mid A) \\
    &= I(M; X) + I(A; X \mid M),
\end{split}
\end{equation}
since $M$ is a random mask independent of $X$, $I(M; X) = 0$. Similarly, we can get $I(A', M ; X) = I(A'; X) + I(M; X \mid A')$. Notice that $A'$ is completely determined by $A$ and $M$, so $I(A,M ; X) = I(A',M ; X)$. It can be obtained:
\begin{equation}
    I(A; X \mid M) = I(A'; X) + I(M;X \mid A').
\end{equation}
Therefore, $I(A; X \mid M) \geq I(A'; X) $. According to the mathematical definition of information, it follows that $I(A; X) \geq I(A; X \mid M) $, and finally, $I(A; X) \geq I(A'; X) $ can be obtained. This leads to the fact that the information bottleneck can support randomized attention masks.

The model results derived after pre-training through parameter sharing are used for downstream fine-tuning, and the multi-view fusion features $F$ are gained by the encoder in Eq.~\eqref{eq:calculateF}. To adapt to the different downstream tasks, we use a different classification head for each task fine-tuning. Thanks to the deep information of CRIA pre-training methodology mining and multi-view fusion of model structure, even with a very simple decoder, CRIA can have outstanding classification performance:
\begin{equation}
    Z = \text{ELU}\left(\text{FC}_1\left(\text{Dropout}(F)\right)\right),
\end{equation}
\begin{equation}
    \mathbf{y^\prime} = \text{ELU}\left(\text{FC}_2\left( \text{LayerNorm}\left(Z\right)\right)\right),
\end{equation}
where $\text{FC}$ represents the Fully Connected layer, $\text{ELU}$\cite{clevert2015fast} is the activation function. $Z$ and $\mathbf{y^\prime}$ are intermediate feature variables and classification results.

For downstream binary classification datasets like TUAB, where the number of classifications is relatively even, fine-tuning uses binary cross-entropy loss (BCE). For unbalanced binary classification datasets similar to CHB-MIT, the focal loss \cite{lin2017focal} is used. In addition, for multicategorization tasks like TUEV, we use multi-class cross-entropy loss\cite{zhang2018generalized}.

\begin{figure}[t]
\centering
\includegraphics[width=0.6\columnwidth]{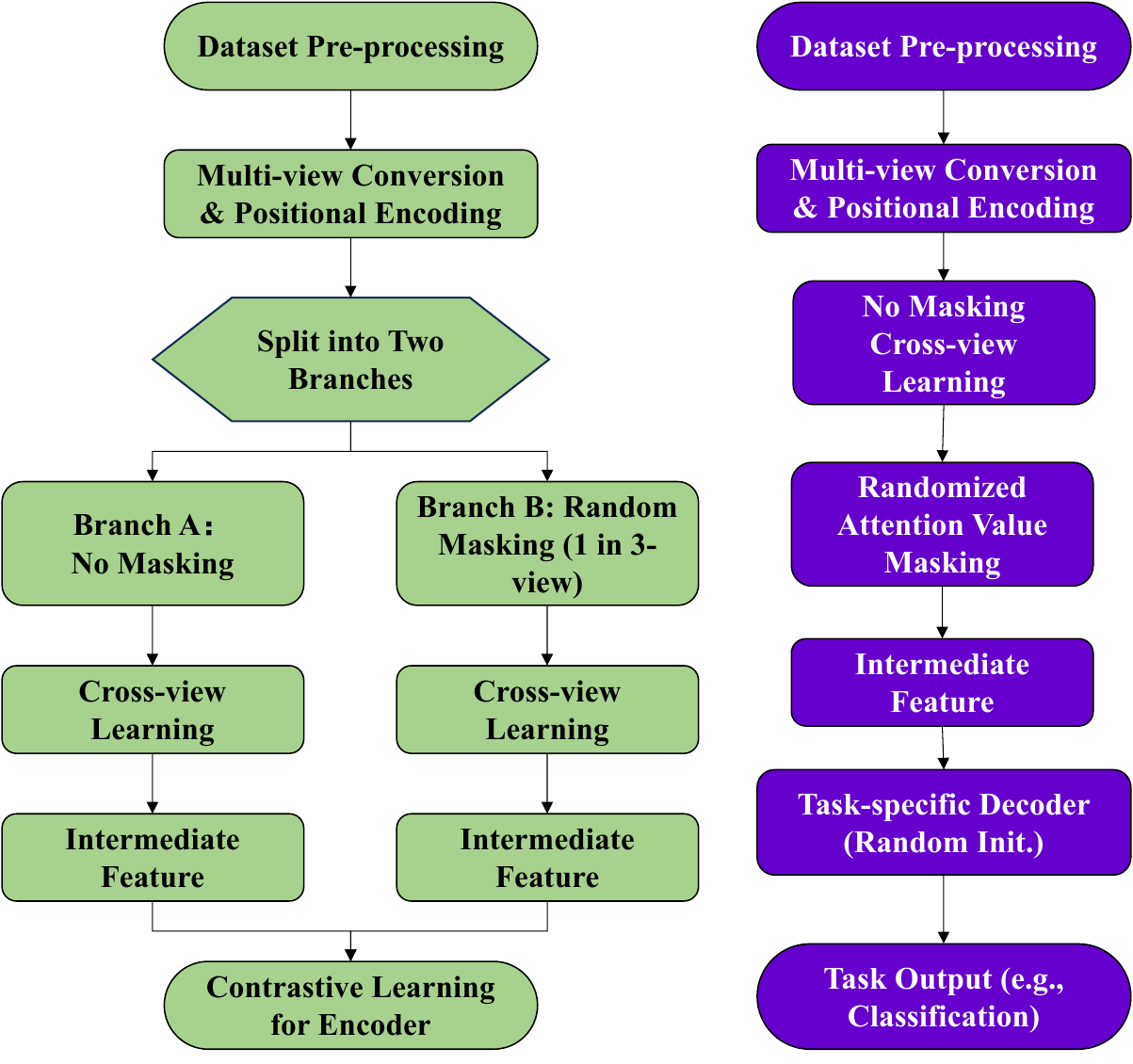} 
\caption{Flowchart of pre-training and downstream fine-tuning.}
\label{TrainingFlowChart}
\end{figure}

As shown in Fig.~\ref{TrainingFlowChart}, pretraining starts with dataset preprocessing, multi-view conversion, and positional encoding. The data is split into two branches: one without masking, processed through a Cross-view Learning module; the other with random masking across views to generate a complementary representation. Both are contrasted via contrastive learning to achieve robust, view-consistent features. During fine-tuning, the same preprocessing and encoding are applied, but without masking. The full multi-view input is passed through the pretrained Cross-view Learning module, and the resulting features are decoded by a randomly initialized task-specific decoder for downstream tasks like classification.

\subsection{Dataset Descriptions}
To ensure that the training of the model is based on publicly available real-world datasets, we used the Temple University EEG Corpus (TUH Corpus)\footnote{https://isip.piconepress.com/projects/tuh\_eeg/
} 
\cite{7002953} for pre-training, and CHB-MIT dataset\footnote{https://physionet.org/content/chbmit/1.0.0/}\cite{guttag10chb,shoeb2009application,goldberger2000physiobank} was used to validate the transferability of the pre-training. TUH Corpus contains several scenario-specific sub-datasets (TUAB, TUAR, TUEP, TUEV, TUSZ and TUSL). Finally, the downstream validation dataset uses TUAB, TUEV and CHB-MIT, which are binary and multiclassification supervised tasks. To ensure stability during pre-training, each batch of data is randomly sampled from a dataset that has been disrupted. Details of the dataset are presented in supplementary material IV.

\subsection{Compared Baselines and Experimental Setups}
The following classes of models are selected for our CRIA comparison experiments:
\begin{enumerate}
    \item Traditional Signal Processing Analyzing CNN Networks: (a) \textbf{SPaRCNet} \cite{jing2023development} is a model designed to automatically classify seizures (SZs) and SZ-like brain activity patterns on EEG with 1D-CNN structure. (b) \textbf{ContraWR}’s \cite{yang2021self} model is a 2D-CNN model developed to automate the analysis of raw EEG data in sleep medicine. (c) \textbf{FFCL} \cite{li2022motor} is a combined CNN and LSTM neural network feature fusion model that considers multiple views simultaneously like CRIA. 
    \item Attention Neural Network: (d) \textbf{CNN-Transformer} \cite{peh2022transformer} is a model that uses CNN for feature extraction and then a Transformer Encoder for automatic artifact detection. (e) \textbf{ST-Transformer} \cite{song2021transformer} is an attention model that focuses on both temporal and spatial view.
    \item EEG Pre-training Model: (f) \textbf{BrainBert} \cite{wang2023brainbert} is an early type of pre-trained neural network that modeled data multi-views like CRIA. (g) \textbf{BIOT} \cite{yang2024biot} is an early unified paradigm that combines pre-training and EEG data to learn the embeddings of biosignals with varying lengths, channels, and missing values. (h) \textbf{EEGFormer} \cite{chen2024eegformer} uses Vector-Quantized Pre-train to characterize EEG in a discrete way. (i) \textbf{EEG2Rep} \cite{foumani2024eeg2rep} utilizes the Semantic Subsequence Preservation (SSP) method to guide the generation of rich semantic representations. (j) \textbf{LaBraM} \cite{jiang2024large} LaBraM uses the VQVAE \cite{van2017neural} approach to quantize EEG data with large-scale cross-dataset coding pre-training. 
\end{enumerate}

To ensure the fairness of the evaluation, we used publicly available code for the baseline methodology. All experiments are implemented in python 3.9.17, Torch 2.0.1 + Cuda 12.2, Pytorch-lightning 2.0.0, and run on a Linux server equipped with 150 GB of RAM, a 64-core CPU, and 3 NVIDIA A100 GPUs. All the models are experimented with using the same data set division. The models are trained on the training set, the best model and hyperparameter combinations are selected based on the validation set, and finally, the test set performance is compared. In the TABLE~\ref{main_results} to~\ref{TrainScratch} and Fig.~\ref{CrossviewAblation} to~\ref{pretrainAblation}, we used five different random seeds for validation and show the mean and standard deviation values. Five layers, including cross-attention, were used for pre-training; the full number of layers was used for downstream validation on the TUAB and CHB-MIT datasets, and only the last three layers close to the task layer were used on the TUEV dataset. Since the pre-training dataset of LaBraM \cite{jiang2024large} is not fully public, to ensure the completeness and fairness of the experiment, we re-pretrained LaBraM using the same pre-training datasets and the same scale as CRIA(ours). The data in the experimental results comes from the downstream tasks we accomplished using the re-pretrained models, and the tuning strategies of the individual models remain consistent and are all tuned to the optimal superparameter combination. This comparison method can effectively avoid the task performance improvement due to data interfering with the comparison of model performance. All experiments passed t significance test to ensure that the performance improvement was not a fluke. All hyperparameters will be shown in TABLE A1 to A4 in the supplementary material I. We will make our code publicly available.

\subsection{Evaluation Metrics}
For the TUAB binary classification task, we adopt balanced accuracy (BACC), the area under the receiver operating characteristic curve (AUROC), and the area under the precision-recall curve (PRAUC) for evaluation. These metrics provide a comprehensive assessment of the model's performance, capturing both the overall accuracy and the balance between precision and recall, as well as the trade-offs between true positive rates and false positive rates.

For the TUEV multi-class classification task, we utilize balanced accuracy (BACC), Cohen's kappa (K), and the weighted F1 score (F1\_Weighted) for evaluation. Balanced accuracy accounts for class distribution imbalances by averaging the recall obtained for each class. Cohen's kappa measures the agreement between the predicted and true classifications, adjusted for chance. The weighted F1 score provides a single metric that balances precision and recall across all classes, weighted by the number of true instances for each class, ensuring robust evaluation against class imbalances.

\section{Experiments}
\begin{table*}
\centering
\small
\caption{For the overall comparison experiments on TUAB and TUEV, ``---'' indicates that this part of the experimental results is not mentioned in the original paper or there is no published pre-training weights. ``*'' represents the multi-classification task. ``S'' for spatial view, ``T'' for temporal view, ``P'' for spectral view. }
\label{main_results}
\resizebox{\textwidth}{!}{%
\begin{tabular}{lcccccccc}
\toprule
\multirow{2}{*}{Model} & \multirow{2}{*}{Views} & \multirow{2}{*}{Pre-train} & \multicolumn{3}{c}{TUAB} & \multicolumn{3}{c}{TUEV*} \\
\cmidrule(lr){4-6} \cmidrule(lr){7-9}
& & & BACC & AUROC & PR AUC & BACC & K & F1 Weighted \\
\midrule
SPaRCNet & T & No & 0.7896 $\pm$ .0018 & 0.8676 $\pm$ .0012 & 0.8414 $\pm$ .0018 & 0.4161 $\pm$ .0262 & 0.4233 $\pm$ .0181 & 0.7024 $\pm$ .0104 \\
ContraWR & ST & No & 0.7746 $\pm$ .0041 & 0.8456 $\pm$ .0074 & 0.8421 $\pm$ .0104 & 0.4384 $\pm$ .0349 & 0.3912 $\pm$ .0237 & 0.6893 $\pm$ .0136 \\
FFCL & ST & No & 0.7848 $\pm$ .0038 & 0.8569 $\pm$ .0051 & 0.8448 $\pm$ .0065 & 0.3979 $\pm$ .0228 & 0.3732 $\pm$ .0188 & 0.6783 $\pm$ .0120 \\
CNN-Transformer & T & No & 0.7777 $\pm$ .0022 & 0.8461 $\pm$ .0013 & 0.8433 $\pm$ .0039 & 0.4087 $\pm$ .0161 & 0.3815 $\pm$ .0134 & 0.6854 $\pm$ .0293 \\
ST-Transformer & ST & No & 0.7966 $\pm$ .0023 & 0.8707 $\pm$ .0019 & 0.8521 $\pm$ .0026 & 0.3984 $\pm$ .0228 & 0.3765 $\pm$ .0306 & 0.6823 $\pm$ .0190 \\
BrainBERT & TP & Yes & --- & 0.8530 $\pm$ .0020 & 0.8460 $\pm$ .0030 & --- & --- & --- \\
BIOT & P & Yes & 0.7959 $\pm$ .0057 & 0.8815 $\pm$ .0043 & 0.8792 $\pm$ .0023 & 0.5281 $\pm$ .0225 & 0.5273 $\pm$ .0249 & 0.7492 $\pm$ .0082 \\
EEGFORMER & P & Yes & --- & 0.8760 $\pm$ .0030 & 0.8720 $\pm$ .0010 & --- & --- & --- \\
EEG2Rep & ST & Yes & --- & \textbf{0.8843} $\pm$ \textbf{.0309} & --- & 0.5295 $\pm$ .0158 & --- & 0.7508 $\pm$ .0121 \\
LaBraM & P & Yes & 0.7903 $\pm$ .0107 & 0.8772 $\pm$ .0123 & 0.8661 $\pm$ .0050 & 0.5626 $\pm$ .0046 & 0.4830 $\pm$ .0145 & 0.7466 $\pm$ .0063 \\
\midrule
CRIA(ours) & STP & Yes & \textbf{0.8003} $\pm$ \textbf{.0012} & 0.8809 $\pm$ .0026 & \textbf{0.8803} $\pm$ \textbf{.0064} & \textbf{0.5702} $\pm$ \textbf{.0324} & \textbf{0.5306} $\pm$ \textbf{.0084} & \textbf{0.7569} $\pm$ \textbf{.0071} \\
\bottomrule
\end{tabular}%
}
\end{table*}

\subsection{Comparison with State-of-the-Art}
\begin{figure}[t]
\centering
\includegraphics[width=0.6\columnwidth]{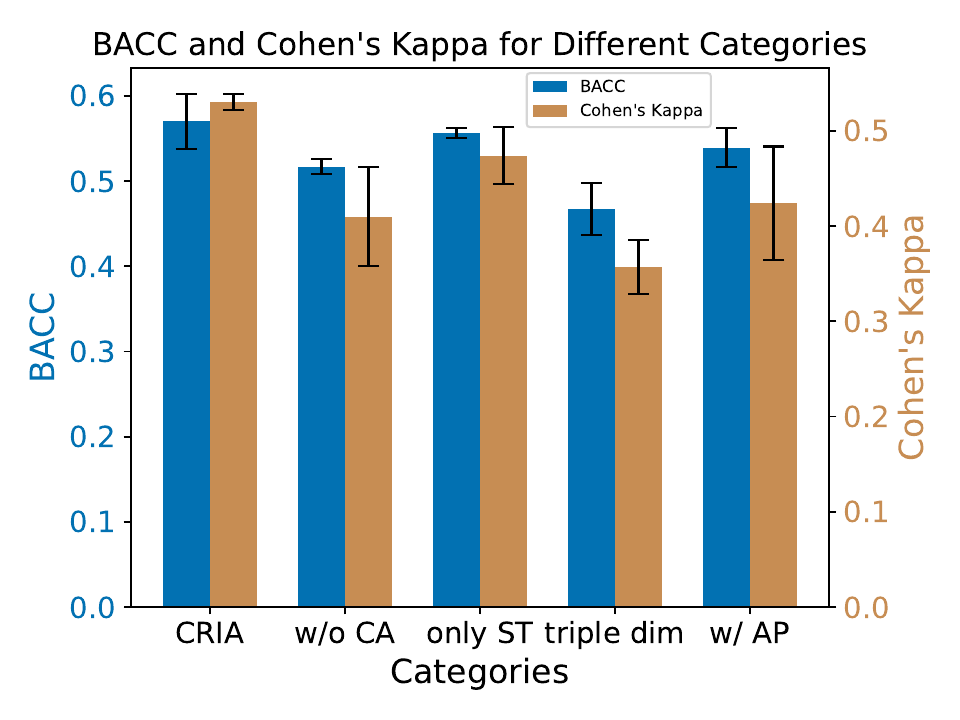} 
\caption{Ablation validation for CRIA in several strategies on TUEV.}
\label{CrossviewAblation}
\end{figure}
The experimental results of the CRIA model and all comparison models on the TUAB and TUEV datasets are presented in the TABLE~\ref{main_results}. All data are obtained by combining the average performance over multiple runs. Bold numbers on each metric indicate the highest accuracy achieved. By observing the results in the TABLE~\ref{main_results}, it can be inferred that the CRIA model achieved the highest average performance in all downstream validation tasks. Models that have been pre-trained with self-supervision tend to achieve better results than other models. While among the non-pre-trained models, ST-Transformer also achieved higher results due to its consideration of multi-view information and fusion of EEG data. In both downstream tasks, TUAB is more adequate and balanced, while the TUEV dataset is highly heterogeneous in terms of categories and has a smaller number of subjects. The pre-trained LaBraM \cite{jiang2024large} employs the VQVAE structure to discretize the EEG data, thereby mitigating the issues of high noise and unevenness in the EEG data, and achieving high performance on downstream tasks. However, due to data discrepancies caused by the non-disclosure of the pre-training dataset, the results in the original article are not achieved. We analyze that most of the huge performance improvement in the original paper may come from the quality and quantity of the private data used for pre-training. 

The CRIA model achieves the best results in the comparison test primarily due to its unique multi-view feature fusion and view-wise masking strategy. First, the CRIA model effectively captures deep multi-view features in EEG data through self-supervised learning, which provides a powerful feature representation capability for its performance in various downstream tasks. Compared to other models, CRIA is able to learn richer signal features without supervised signals, making it more adaptable to unseen data and exhibiting better generalization ability. In addition, CRIA is able to synthesize signal features from different channels and time periods, and this multidimensional feature aggregation enables the model to maintain stable performance better when facing complex and highly heterogeneous tasks.

\begin{table}[h]
\centering
\caption{Results of classification experiments migrated to the CHBMIT dataset after pre-training on the TUH Corpus dataset.}
\label{TransToCHBMIT}
\scalebox{0.7}{ 
\begin{tabular}{lccccc}
\toprule
\multirow{1}{*}{Model} & \multicolumn{3}{c}{From TUH Corpus to CHB-MIT}\\
\cmidrule(lr){2-4} 
& BACC & PR AUC & AUROC \\
\midrule
SPaRCNet & 0.5876 $\pm$ .0191 & 0.1247 $\pm$ .0119 & 0.8143 $\pm$ .0148 \\
ContraWR & 0.6344 $\pm$ .0002 & 0.2264 $\pm$ .0174 & 0.8097 $\pm$ .0114 \\
CNN-Transformer & 0.6389 $\pm$ .0067 & 0.2479 $\pm$ .0227 & 0.8662 $\pm$ .0082 \\
FFCL & 0.6262 $\pm$ .0104 & 0.2049 $\pm$ .0346 & 0.8271 $\pm$ .0051 \\
ST-Transformer & 0.5915 $\pm$ .0195 & 0.1422 $\pm$ .0094 & 0.8237 $\pm$ .0491 \\
BIOT & 0.7068 $\pm$ .0457 & 0.3277 $\pm$ .0460 & 0.8761 $\pm$ .0284 \\
LaBraM & 0.7980 $\pm$ .0025 & \textbf{0.4007} $\pm$ \textbf{.0150} & 0.8568 $\pm$ .0027 \\
\midrule
CRIA(ours) & \textbf{0.8379} $\pm$ \textbf{.0186} & 0.3982 $\pm$ .0164 & \textbf{0.8770} $\pm$ \textbf{.0066} \\
\bottomrule
\end{tabular}%
}
\end{table}

\subsection{Pre-training Migration Across Datasets}
In order to verify the migrability of the pre-trained model, we pre-trained CRIA on the TUH Corpus dataset and then completed the classification task on the CHB-MIT unseen dataset. According to the TABLE~\ref{TransToCHBMIT}, CRIA is able to migrate to the unseen dataset effectively, demonstrating good generalization ability. Through self-supervised pre-training on multiple datasets, the CRIA model learns a fused in-depth representation of the multi-view features of the EEG data, which enables it to maintain high performance in the face of new, unseen data. Especially when migrating to a more uneven unseen dataset such as CHB-MIT, the CRIA model successfully adapts to the new data distribution and category structure, with its advantage in fusion of multi-view information, far surpassing other models that are not pre-trained. Pre-training not only improves the model's ability to understand different features, but also enhances the consistency of its performance in different tasks, proving the strong adaptability and robustness of the CRIA model in multi-scenario applications.

\begin{figure*}[t]
\centering
\includegraphics[width=1\textwidth]{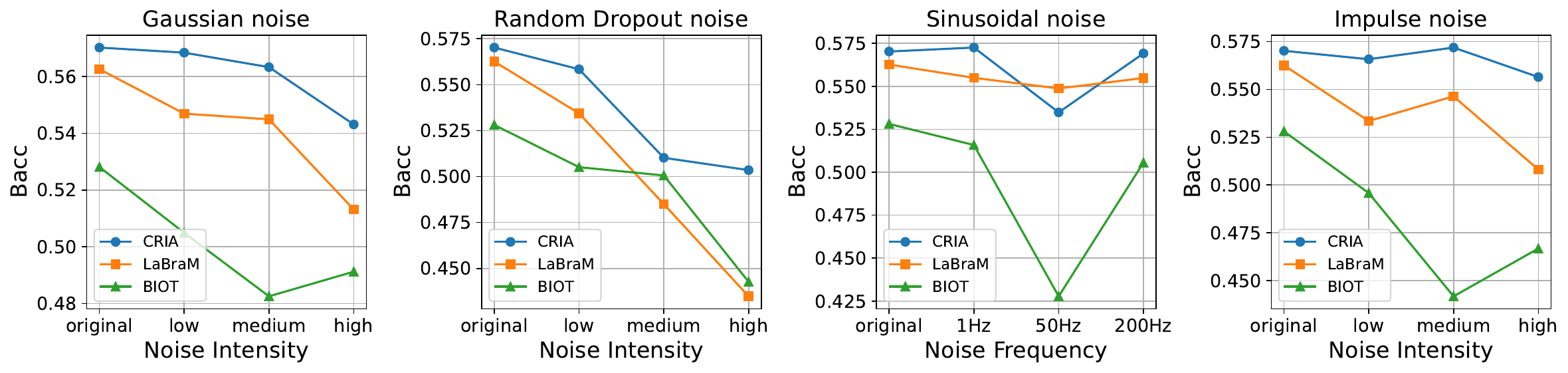}
\caption{Model Robustness Experimental Results. Adding different types of noise to the TUEV dataset to verify the effect on the model results.}
\label{Robustness}
\end{figure*}

\subsection{Training from Scratch}
The results in TABLE~\ref{TrainScratch} show that CRIA achieves superior performance even without the aid of pre-training compared to other models that use random initialization. Specifically, on the TUAB dataset, CRIA achieves the highest AUROC (0.875) and PR AUC (0.871), which are significantly better than other competing models. On the TUEV dataset, CRIA outperforms other methods in both BACC (0.545) and F1 weighting (0.735), demonstrating its ability to handle downstream tasks directly from scratch. This performance improvement is attributed to CRIA's unique design, which emphasizes learning from multiple views simultaneously.

Unlike traditional approaches that rely excessively on time-consuming and computationally intensive pre-training processes and large, high-quality datasets, CRIA utilizes mutual information augmentation across views, enabling the model to better capture both shared and distinct features across views in non-pre-training situations. This focus on multi-view consistency greatly improves the model's generalization ability and representation quality, even in the absence of pre-training. By eliminating the reliance on pre-training and maintaining high accuracy, CRIA sets a new benchmark for pre-training-free methods in EEG analysis. This capability makes CRIA a promising approach for real-world applications where time efficiency and computational cost are severely limited.

\begin{table}[h]
\centering
\caption{Pre-training-free randomized parametric experiments on TUAB and TUEV.}
\label{TrainScratch}
\scalebox{0.7}{ 
\begin{tabular}{lcccccc}
\toprule
\multirow{1}{*}{Model} & \multicolumn{2}{c}{TUAB} & \multicolumn{2}{c}{TUEV*} \\
\cmidrule(lr){2-3} \cmidrule(lr){4-5}
& AUROC & PR AUC & BACC & F1 Weighted \\
\midrule
BIOT & 0.869 $\pm$ .003 & 0.871 $\pm$ .009 & 0.468 $\pm$ .013 & 0.708 $\pm$ .018 \\
EEG2Rep & 0.849 $\pm$ .030 & --- & 0.443 $\pm$ .031 & 0.689 $\pm$ .028 \\
LaBraM & 0.824 $\pm$ .054 & 0.810 $\pm$ .024 & 0.522 $\pm$ .009 & 0.721 $\pm$ .036 \\
\midrule
CRIA(ours) & \textbf{0.875} $\pm$ \textbf{.005} & \textbf{0.871} $\pm$ \textbf{.004} & \textbf{0.545} $\pm$ \textbf{.014} & \textbf{0.735} $\pm$ \textbf{.018} \\
\bottomrule
\end{tabular}%
}
\end{table}

\subsection{Ablation Validation}
\subsubsection{Influence of Multiple Views and Mutual Feature}
To validate our method of focusing on unique information from three EEG views while capturing inter-view features using cross-attention, we conducted ablation studies, as shown in Fig.~\ref{CrossviewAblation}. We compared full CRIA against variants: self-attention replacing cross-attention (w/o CA), using only time and space views (only ST), and tripling attention dimensionality for frequency-domain features (triple dim).

The full CRIA model performed best, with significant improvements in both BACC and Cohen's Kappa for CRIA compared to w/o CA. This suggests that focusing on three-view information and capturing complementary features across views through the cross-attention mechanism is key to model performance improvement. only ST performs slightly lower than CRIA, suggesting that the introduction of frequency-domain features can provide important complementary information that may be critical for categorizing an event or a specific task. Triple dim, although increasing the dimensionality of frequency-domain features, does not significantly improve performance. This may be due to the fact that simply increasing the dimensionality does not adequately model the deep semantic relationships of the frequency domain information, but instead introduces redundant features, and also highlights the importance of parallel cross-attention in three views.
\subsubsection{Contribution of Representation Enhanced Purification}
The Fig.~\ref{CrossviewAblation} also shows the strategy of automatically selecting channels and time slices before fusion using the attention mechanism approach (in CRIA) and the strategy of directly using average pooling fusion (w/AP). The full CRIA shows higher performance on both BACC and Cohen's Kappa, which suggests that it is better able to localize event-related channels and time segments in the EEG data. The w/AP method, on the other hand, simply averages all features, ignoring the differences in the importance of different channels and temporal segments to the classification task, which leads to underutilization of information. The representation enhanced purification strategy improves the ability to focus on the key features of the EEG data by automatically selecting on the temporal and channel dimensions, resulting in a more stable and superior performance of the classification task.

\subsection{Model Robustness Validation}
In order to verify the robustness of the model, we added different types of noise to the TUEV dataset as a way to test its effect on the experimental results in Fig.~\ref{Robustness}. Detailed noise setups and experimental results are presented in supplementary material II.

According to the experimental results, CRIA generally exhibits high robustness in various types and intensities of noise environments. When the intensity of Gaussian and Impulse noise varies from low to high, CRIA's performance (BACC) decreases significantly less than that of LaBraM and BIOT. This indicates that CRIA is able to cope more efficiently with random high-frequency disturbances in real-world situations, as well as sudden short-term large-value variations in, for example, device currents. This is due to the fact that CRIA extracts features of the EEG data from different views. And while receiving interference at a certain frequency, a voting mechanism can still be used to find the special diagnostic components that favor the downstream task.

CRIA outperforms other methods under random dropout noise, achieving higher BACC than LaBraM and BIOT at all noise intensities, especially at high intensity noise where it still maintains a high classification accuracy. LaBraM's two-stage pre-training with VQVAE may reduce robustness to dropout noise, while it remains reliable against 50 Hz sinusoidal noise, as its pre-training captures this regular frequency component. VQVAE is able to ignore regular small-amplitude perturbations in the input data by discretizing the latent-space representation. CRIA versus the training method of a randomly-masked view, which also provides its advantage in the case of random dropout of signal, but like BIOT, there is a performance degradation in the face of sinusoidal noise.

\subsection{Insight of View-wise masking}
\begin{figure}[t]
\centering
\includegraphics[width=0.6\columnwidth]{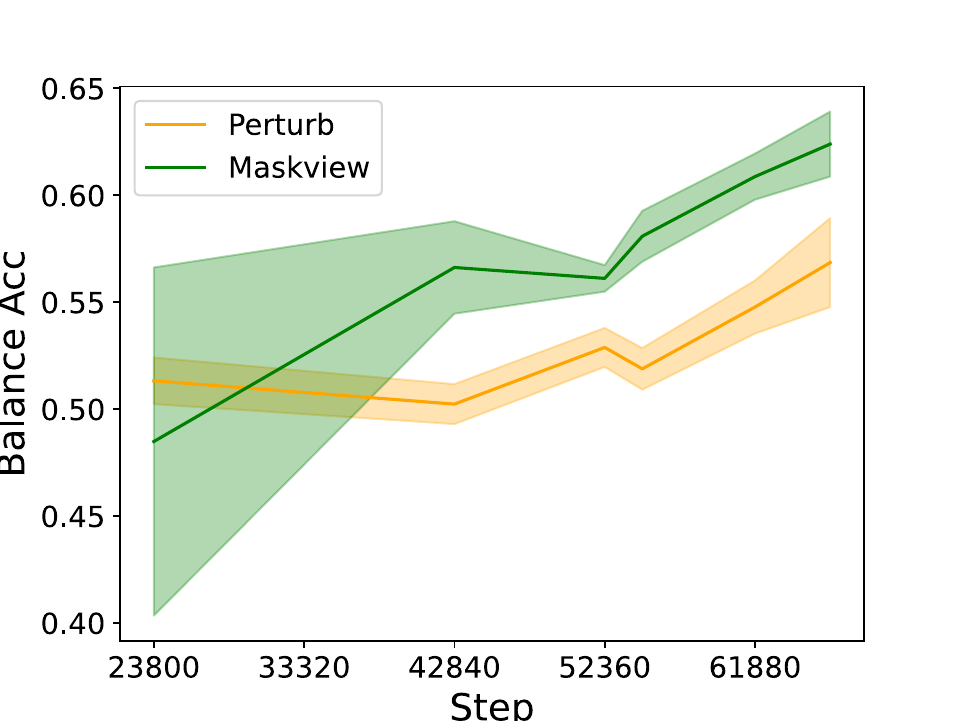} 
\caption{Changes in downstream performance of Perturb and Maskview strategies as the number of pre-trained steps increases on the TUEV dataset.}
\label{pretrainAblation}
\end{figure}

We compare the Perturb pre-training method from BIOT \cite{yang2024biot} with our view-wise mask pre-training method on the same dataset and downstream task (TUEV). Fig.~\ref{pretrainAblation} shows the downstream task performance as pre-training steps increase. The view-wise mask strategy exhibits better performance and faster convergence. Due to EEG's non-stationary nature and low signal-to-noise ratio, an easier pre-training task is beneficial. View-wise masking allows training with one view masked and another shown, preserving some regional information. This approach is akin to a student mastering basic questions rather than struggling with difficult ones, which leads to better test performance.

\section{Conclusion}
In this paper, we propose CRIA, a novel pre-training framework for EEG data analysis that introduces an asymmetric three-view interaction mechanism to achieve structured and semantically-aware fusion across different views. CRIA enhances the feature representation through a cross-attention mechanism, adapts to variable-length and channel EEG data, and achieves cross-dataset and cross-task generalizability. The view-masking pre-training strategy improves model performance and convergence speed, effectively optimizes the pre-training convergence problem caused by the non-stationarity and low signal-to-noise ratio of EEG data. In experiments on the Temple University EEG corpus, CRIA excels in anomaly detection and multi-class event classification, maintaining outstanding performance even when trained from scratch or when encountering different types of noise, thereby demonstrating its robust architecture and feature extraction capabilities. However, the practical effects and potential challenges of CRIA in clinical applications still need to be further explored, including patient specificity, the impact of diverse pathologic conditions on generalizability, and the trade-off between computational efficiency and the interpretability of results in real-time integration. Future work will focus on large-scale pre-training and real-time processing studies to promote the translation of algorithmic innovations into practical clinical applications.

\bibliographystyle{elsarticle-num}
\bibliography{cas-refs}
\end{document}